# The Impact of Foundational Models on Patient-Centric e-Health Systems

Elmira Onagh
*York Univeristy*
eonagh@yorku.ca

Alireza Davoodi
*Curatio.me*
alireza@curatio.me

Maleknaz Nayebi
*York Univeristy*
mnayebi@yorku.ca

*Abstract*—As Artificial Intelligence (AI) becomes increasingly embedded in healthcare technologies, understanding the maturity of AI in patient-centric applications is critical for evaluating its trustworthiness, transparency, and real-world impact. In this study, we investigate the integration and maturity of AI feature integration in 116 patient-centric healthcare applications. Using Large Language Models (LLMs), we extracted key functional features, which are then categorized into different stages of the Gartner AI maturity model. Our results show that over 86.21% of applications remain at the early stages of AI integration, while only 13.79% demonstrate advanced AI integration.

## I. INTRODUCTION

Artificial Intelligence (AI) is rapidly gaining traction across various sectors, including health care. However, the current state and maturity of its integration into real-world mobile health applications remain largely underexplored. In particular, it is not yet clear who the primary users of these AI-enabled features are, patients or health care providers, and for what specific purposes they are being employed. Foundational Models (FMs), large-scale AI models trained on diverse and extensive datasets, have recently emerged as a transformative force across multiple domains. These models, including Large Language Models (LLMs), vision models, and multimodal systems, are increasingly integrated into the sociotechnical landscape due to their capacity to generalize tasks with minimal fine-tuning. Among the many sectors exploring their potential, health care stands out for its growing efforts to harness FMs to support clinical decision-making, streamline workflows, and enhance the quality of patient care.

Within this context, Health Foundation Models (HFMs), a specialized subclass of FMs trained on biomedical texts, Electronic Medical Records (EMRs), and clinical notes, enable applications such as predictive diagnostics, personalized treatments, and natural language interfaces [70]. HFMs hold strong potential for patient-centric e-Health by empowering users, enhancing engagement, and supporting personalized, high-quality care. Despite these opportunities, integrating HFMs into real-world healthcare settings presents significant challenges [15]. Key concerns include data privacy, model transparency, algorithmic bias, regulatory compliance, and the risk of overreliance on AI in high-stakes clinical decisions. For instance, the use of underrepresented data in training can lead to biased outcomes, amplifying health disparities among patients of color. Moreover, aligning these models with patient-centred values, such as trust, interpretability, and inclusivity, remains an ongoing area of research. As such, understanding the current landscape of patient-centric e-Health and the role HFMs may play in shaping its future is both timely and necessary.

App stores, especially the Google Play Store, have shown great potential as a valuable resource for analyzing trends, identifying popular categories, and understanding the competitive landscape in the digital health space [22], [71]. When developing a new mobile application, it is crucial to understand the surrounding ecosystem, such as the app store environment. Mining app store data provides invaluable insights into widely adopted features and emerging services, enabling companies to identify proven functionalities that can be seamlessly integrated into existing applications. Market entry decisions should be informed by comparative analyses of competing applications, both in terms of functionality and the maturity of those features. This process aligns with the growing emphasis on data-driven approaches in software evolution and requirements engineering, where feature reuse plays a key role in accelerating development and improving software reliability [7], [25]. The use of mobile app stores also taps into the concept of multi-sided markets, where diverse stakeholders, including app developers, end-users, and service providers, can collaborate to offer innovative solutions. This dynamic allows companies to leverage services and features from a wide array of developers, reducing costs and enhancing the overall value proposition of their apps [5], [35], [58]. In the context of digital health, such strategies can significantly improve the functionality and user appeal of apps, positioning them as competitive and feature-rich solutions within the rapidly evolving e-Health ecosystem [11], [18], [64].

In this study, we aim to understand the current landscape of patient-centric e-Health and the role of FMs by examining the e-Health solutions available on the Google Play Store. We address the following research questions:

**RQ1:** What is the current status quo of FM integration in e-Health applications?

To address this question, we identified 116 applications on the Google Play Store that focus on patient support. We systematically mined the data for each application and extracted the functionalities using state-of-the-art methods to assess the extent of AI and FMs integration.

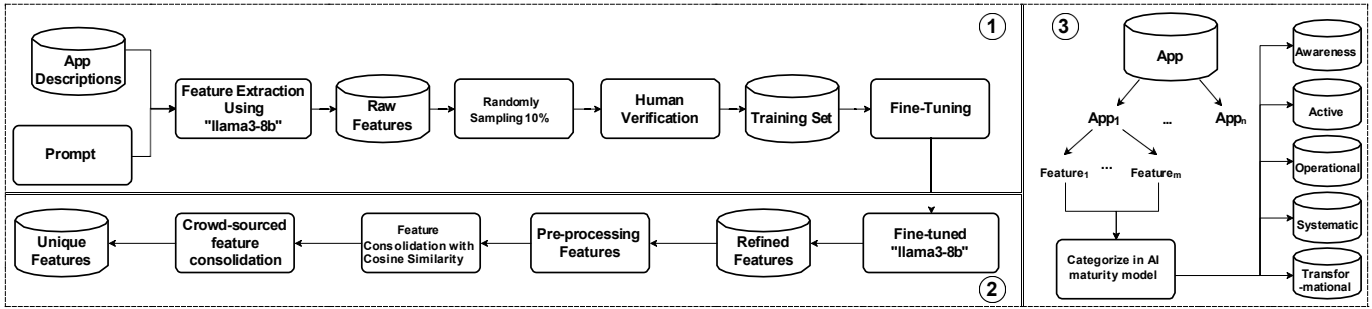

Fig. 1. Methodology Overview: (1) We prepared a dataset to fine-tune the LLaMA model. (2) The model was then used to extract and consolidate application features. (3) Each app was classified into an AI maturity level using the Gartner framework.

**RQ2:** Given the growing popularity of FMs, what is the current state of AI maturity in e-Health applications?

To address this question, we utilize the AI maturity model (AIMM) introduced by Gartner Group [13], a widely adopted framework in both research and practice for assessing the level of AI maturity of products. We categorized the applications in our dataset according to this model and examined the impact of AI maturity on the popularity of the applications.

We review related work in Section II, describe our methodology in Section III, present results in Section IV, and discuss findings in Section V. Section VI outlines validity threats, and Section VII concludes the paper.

## II. RELATED STUDIES

AI has seen widespread adoption in healthcare, particularly in diagnostics, treatment planning, and operational efficiency. Secinaro et al. [62] reviewed 288 multidisciplinary studies, identifying key trends in health service management, diagnostics, and predictive medicine. Similarly, Al Kuwaiti et al. [4] emphasized AI's transformative impact, especially during COVID-19, across virtual care, drug discovery, and administration, highlighting ethical, technical, and governance challenges. Despite focusing on the provider side of the applications, AI's role in patient-centric technologies remains underexplored. This is notable given patients' growing use of digital tools for self-monitoring, education, and early diagnosis. AIMMs have primarily been developed to assess organizational readiness or the technical robustness of AI systems. These models typically evaluate dimensions such as explainability, adaptability, and data handling capabilities and have been extensively studied in academia [6], [13], [23], [23], [52], [72]. However, their application to consumer-facing healthcare tools is limited. Our work addresses this gap by mapping observable features of healthcare applications to an AI maturity spectrum, offering a novel lens for evaluating their technical evolution and potential risks. Prior work has used NLP, heuristics, and crowdsourcing to classify healthcare apps by functionality or compliance [8], [16], [69], [73]. While LLMs show promise in Requirements Engineering (RE) tasks such as feature extraction, testing, etc., and can approximate human judgment [9], [26], [68], [74], challenges like hallucination and lack of transparency persist. We address this by integrating fine-tuned LLMs with crowdsourced validation for more robust AI feature extraction.

## III. METHODOLOGY

To achieve the objectives of this study, we first extracted e-Health-related applications using `google-play-scraper` [17] library. The applications were identified using the following keywords: "Patient engagement", "Healthcare community", "Chronic disease management", "Health management", "Peer support health", "Patient support network", "Health social network", "Digital health engagement", and "Health community app". For each application, we extracted the following metadata: `appId`, `title`, `score`, `genre`, `price`, `description`, `developer`, and `installs`.

To identify key functionalities of the apps (**RQ1**), we leveraged LLMs for feature extraction using app descriptions. Given the growing success of LLMs in RE tasks and their ability to approximate human evaluation [9], [26], [68], [74], we selected the `llama3.1-8b` model after manual comparison across candidates. The model was prompted to extract raw features, and a manual review of 146 randomly selected features revealed a 6.16% inaccuracy rate. These were corrected to improve alignment with app descriptions.

To enhance accuracy and reduce hallucinations, we fine-tuned the model using the corrected samples and re-ran it on the dataset. Since similar features can appear under different phrasings, we normalized and lemmatized the extracted features, converted them into vector embeddings using `bert-base-nli-mean-tokens` [10], and applied cosine similarity (threshold = 0.9 [1]) via `scikit-learn` [61] to cluster and consolidate features. Final refinements were made with human validation through crowd-sourcing. The overview of the feature extraction methodology is depicted in Figure 1. Using these extracted features and discussions with experts in the e-Health application development field, we manually classified each application into one of the levels of AIMMs (**RQ2**). This framework is widely used in the industry to measure the maturity of companies and their products in terms of AI capabilities [54] (more details in Section III-A).

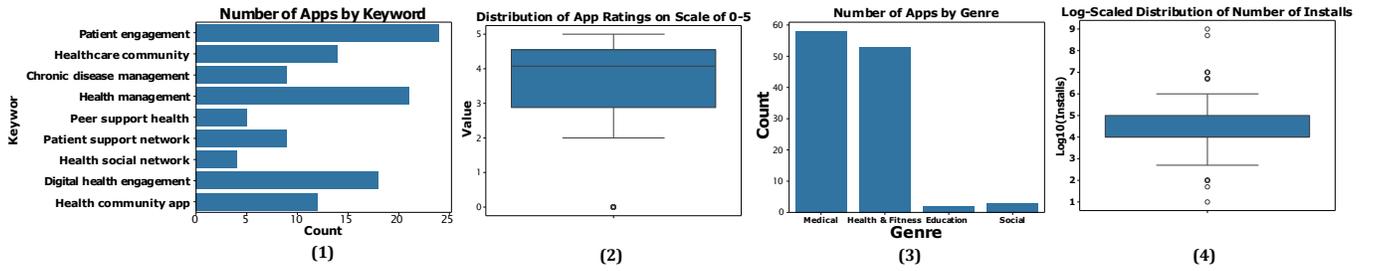

Fig. 2. Analysis patient-centric e-Health apps: (1) keyword distribution, (2) ratings distribution, (3) genre distribution, (4) log-scale install count distribution.

Finally, we performed a comprehensive analysis of the data to assess the impact of FMs on the e-Health landscape, providing insights into how AI influences the development and adoption of e-Health solutions. The replication package used in this study is provided in our GitHub repository [63].

*A. AI Maturity Models (AIMMs)*

With the growing adoption of AI and FMs, organizations increasingly use AIMMs to guide and assess their integration efforts [6]. Among available frameworks, the Gartner AIMMs [13], [23] is widely referenced in both academia and industry [6], [23], [52], [72]. We adopt this model to evaluate the level of AI integration in patient-centric applications. It defines five distinct levels of maturity:

**Awareness**: AI-related discussions occur informally and lack strategic direction, with no pilot initiatives underway.
**Active**: Organizations begin exploring AI through pilot projects or proofs of concept, and preliminary conversations around standardization start to emerge.
**Operational**: At least one AI initiative is in production, backed by dedicated funding, executive support, and internal AI expertise and best practices.
**Systematic**: AI is embedded in the design of new products and services, and employees across departments incorporate AI into processes and applications.
**Transformational**: Although not explicitly defined in detail, this stage implies a comprehensive and strategic integration of AI that fundamentally reshapes organizational operations, driving sustained innovation and competitive advantage.

Gartner reports [13] highlight that most organizations are still in the early stages of AI maturity, with barriers such as identifying high-impact use cases, privacy concerns, integration challenges, and unrealistic timelines.

## IV. Preliminary Results

We identified 186 applications using the keyword-based approach described in Section III. However, upon manual inspection, we found several applications unrelated to healthcare. Broad terminologies such as "community" and "network" led to the inclusion of general-purpose social media platforms, such as Facebook and Reddit. To ensure the relevance of our dataset, we manually excluded these unrelated entries, resulting in a refined set of 116 patient-centric applications that offer various forms of support to patients. Figure 2-1 illustrates the distribution of collected applications based on the keywords used during extraction.

Applications on the Google Play Store can be either free or paid. In our dataset, only one application required payment to install; the rest were free to download, although many included in-app purchases or required referral codes for access to specific features. App ratings and install counts are widely used as proxies for popularity [12], [57], [67]. The average rating across the applications was 3.32, on a scale from 0 to 5, with the distribution shown in Figure 2-2.

Given the wide variance in install counts, we applied a logarithmic transformation to normalize the data. After this transformation, the average log-scaled install count was 4.53, corresponding to an average of 33,884 installs per application (Figure 2-4). Moreover, the majority of applications in our dataset are categorized under the "Medical" and "Health & Fitness" genres in the Google Play Store, with fewer classified as "Education" and "Social" (Figure 2-3).

*A. Functionalities in e-Health Applications (RQ1)*

In total, we extracted 942 unique features across 116 applications in our dataset using the methodology outlined in Section III. After consolidating these features to remove duplicates and variations in phrasing, we found that the average number of features per application was 11.12. This suggests that, on average, each application offers a moderate range of functionalities tailored to patient support and care.

Figure 3-1 presents a heatmap of the top 10 features and their co-occurrence with other features. The diagonal elements of the heatmap represent the frequency with which the apps in our dataset offer each feature. Among the most popular features, we identified "Provide Health Data" (offered 21 times), "Access Medical Record" (14), "View Medication" (10), "Symptom Tracking" (9), and "Schedule Upcoming Appointment" (9). These features stand out as essential components of patient-centric e-Health solutions.

Additionally, our co-occurrence analysis revealed interesting patterns of feature interactions. For instance, "Provide Health Data" showed a weak co-occurrence with both "Symptom Tracking" and "View Medication", suggesting that while the apps commonly offer these features, they are not often combined in the same application. Similarly, "Symptom Tracking" exhibited weak co-occurrence with "Mood Tracking", indicating that while both are related to monitoring patient well-being, they are not frequently integrated in the same app. These findings provide valuable insights into the feature landscape of patient-centric e-Health applications and their design patterns.

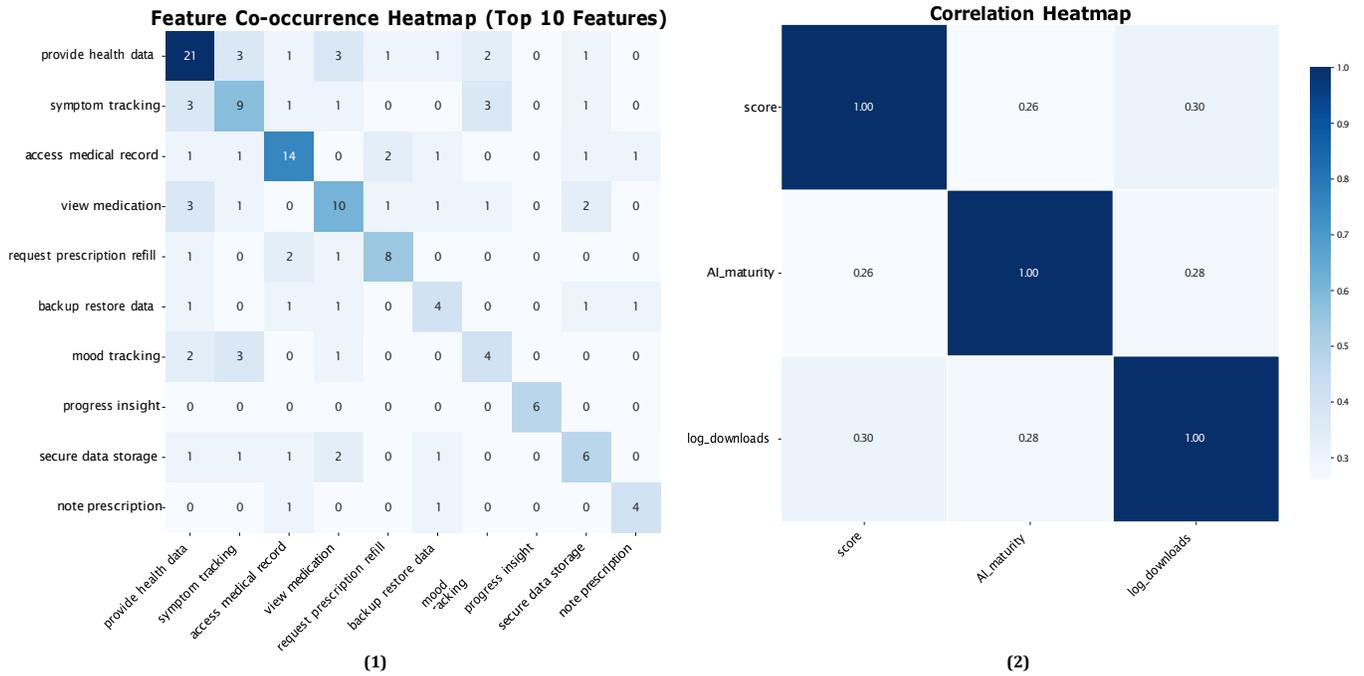

Fig. 3. Co-occurrence analysis of patient-centric apps: (1) top 10 feature co-occurrence (diagonal lines show feature frequency), and (2) correlation matrix between AI maturity, installs, and ratings.

### B. Extent of AI Maturity in Digital Health landscape (RQ2)

We manually categorized each application using the extracted features in our dataset based on the levels defined in the AIMM (Section III-A), which reflects the extent to which AI is integrated into an application's core functionalities. Among the 116 applications analyzed, 70 were classified in the "Awareness" stage, indicating minimal or superficial use of AI, such as basic rule-based systems or static information delivery. This was followed by 30 applications in the "Active" stage, where AI functionalities are present but limited in scope, and 11 in the "Operational" stage, representing more consistent and interactive AI features. Only five applications reached the "Systematic" stage, indicating a deeper and more integrated use of AI across multiple aspects of the application. Notably, no applications were found in the "Transformational" stage, which would indicate a paradigm shift enabled by AI, such as personalized treatment recommendations or autonomous health decision support. To examine the potential relationship between AI maturity and user perception, we plotted the distribution of applications across AI maturity levels alongside their average user ratings (Figure 4). Although a higher AI maturity level does not necessarily guarantee higher user satisfaction, the data indicate a positive trend: Applications in higher AI maturity generally receive better average ratings than those in lower tiers. Additionally, we calculated the correlation between AI maturity levels and app popularity metrics, including user ratings and number of downloads (Figure 3-2). Our analysis revealed a weak but positive correlation (0.30) with both metrics, suggesting that applications may experience positive gains in user engagement and perceived value as AI maturity increases.

## V. DISCUSSION

Our analysis of 116 patient-centric e-Health apps provides key insights into AI integration and maturity. We will discuss our findings in the following section. Our analysis of 116 patient-centric e-health apps provides key insights into AI integration and maturity. We will discuss our findings in the following section. This work builds upon the existing body of literature developed over the years [2], [3], [3], [14], [19]–[21], [24], [27], [28], [28]–[45], [45]–[51], [53], [55], [56], [58], [59], [59], [60], [65], [65], [66], [66].

### A. Integration of AI & FMs in e-Health Applications (RQ1)

The feature extraction process revealed that while AI functionalities are present in many patient support apps, they are often limited in scope and depth. Most applications rely on basic automation, chatbots, or rule-based personalization features, suggesting that AI integration is still predominantly at the surface level. Advanced functionalities leveraging deep learning or FMs, such as predictive analytics, intelligent recommendations, or context-aware feedback, remain relatively rare. This gap indicates that the full potential of FMs in e-Health is yet to be realized, particularly in patient-facing solutions. Additionally, many commonly recurring features (e.g., providing health data, symptom tracking, accessing medical records) do not require high-level AI capabilities but indicate strong UX practices and user-centric design. This suggests that, at present, the value proposition of many e-Health apps lies more in usability and accessibility than in sophisticated AI-driven functionality.

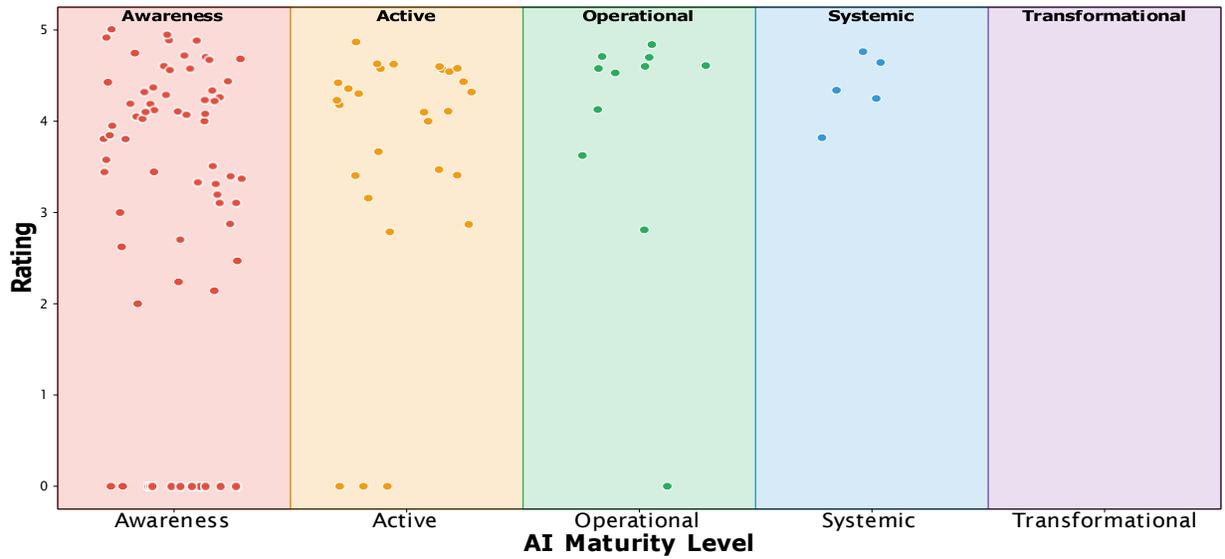

Fig. 4. Categorization of patient-centric e-Health applications by AI maturity level and user rating distribution.

### B. AI Maturity and Its Impact (RQ2)

The classification of applications using the AIMMs revealed that many apps fall within the *Awareness* or *Active* stages. Only a small fraction have reached the *Operational* or *Systematic* levels, with none qualifying as *Transformational*. This finding supports broader observations in the literature [6], [13] that healthcare organizations often struggle to scale AI initiatives beyond isolated use cases. One important implication is that although the e-Health ecosystem is rich in experimentation, it lacks systemic integration of robust, scalable AI capabilities aligned with clinical workflows. This fragmented maturity also limits the ability of HFMs to deliver on their promise in patient-centric contexts, as their deployment often requires a sophisticated digital infrastructure and trust-enabling mechanisms, such as explainability and compliance frameworks. Moreover, we observed a positive correlation between higher AI maturity levels and indicators of popularity (e.g., higher ratings or install counts), hinting at a market preference for more intelligent, responsive, and context-aware applications. However, causality cannot be conclusively established, and further studies are required to examine whether AI maturity directly contributes to user retention and app success or if both are consequences of more capable development teams, better funding, or any other factors..

### C. Implications for Industry and Research

From an industry perspective, our findings highlight an opportunity for developers and healthcare providers to more strategically adopt AI capabilities, especially those enabled by FMs. Rather than focusing solely on technological novelty, developers should prioritize functionalities that demonstrably improve patient engagement, care coordination, and long-term health outcomes. As demonstrated in this study, reusing proven functionalities via app store mining can accelerate the development of robust, AI-enhanced features without reinventing the wheel. For researchers, this study underlines the need to bridge the gap between FM innovation and real-world implementation in patient-facing tools. Current research in HFMs tends to prioritize model performance on benchmark datasets; however, practical deployment in e-Health settings requires a concurrent focus on usability, safety, interpretability, and alignment with regulatory and ethical frameworks. Future work may explore more nuanced maturity models tailored to healthcare applications, incorporating clinical validation, user trust, and interoperability dimensions.

## VI. THREATS TO VALIDITY

This section discusses threats to the validity of our study. Despite careful execution, our findings may be influenced by the following:

**Construct Validity:** Construct validity concerns whether our study accurately captures the concept of AI maturity and functional integration. To ensure the validity of our study, we built on established frameworks and iteratively refined our schema through team discussions. However, classifying AI capabilities and maturity levels inevitably involves subjective judgment, which may have introduced bias.

**Internal Validity:** Our results rely on observable features extracted from application descriptions. However, some applications may include undocumented backend AI functionalities. Additionally, using LLMs for feature extraction poses a risk of hallucination. To address these issues, we developed a domain-informed keyword set in consultation with experts to identify patient-centric applications. We used a fine-tuned model for extraction and manually validated the results through crowd-sourced evaluation.

**External Validity:** External validity pertains to the generalizability of our findings. Our study focused on a curated sample of 116 patient-centric e-Health applications from a major marketplace, which may not represent other health technologies. However, our dataset's diverse functionality and geographic coverage clearly represent current market practices.

**Conclusion Validity:** Conclusion validity addresses the reliability and robustness of our interpretations. We refrain from making causal claims while we report descriptive statistics and comparative analyses. Our categorization process, though systematic, is based on qualitative judgments and may be subject to coder bias. To mitigate this, we used team-based coding, maintained detailed decision logs, and reached a consensus on ambiguous cases.

## VII. Conclusion

This study offers a first-of-its-kind analysis of AI maturity and functional integration across 116 patient-centric e-Health applications. Despite growing interest in AI, especially FMs, adoption remains limited and often superficial. Most apps exhibit low AI maturity, with few employing advanced features like personalized insights or conversational support. However, early signs point to a shift toward more context-aware, patient-centred AI. As the field evolves, collaboration among developers, researchers, and regulators is essential to ensure transparent, safe, and clinically relevant AI use. This work provides a baseline for tracking the real-world impact of FMs in e-Health.